\documentclass[letterpaper, 10 pt, conference]{ieeeconf}
\usepackage{graphicx}
\usepackage{multirow}
\usepackage{hyperref}
\usepackage[percent]{overpic}
\usepackage{tikz}
\usepackage{amsmath}

\IEEEoverridecommandlockouts
\title{\LARGE \bf
  Self-Modifying Morphology Experiments with DyRET:\\ Dynamic Robot for Embodied Testing*
}

\author{T{\o}nnes F. Nygaard$^{1}$, Charles P. Martin$^{1}$, Jim Torresen$^{1,2}$ and Kyrre Glette$^{1,2}$
\thanks{*This work is partially supported by The Research Council
  of Norway as a part of the Engineering Predictability with
  Embodied Cognition (EPEC) project, under grant agreement
  240862, and through its Centres of Excellence scheme, project number 262762.}
\thanks{Authors are with $^{1}$the Department of Informatics, and $^{2}$RITMO,
        University of Oslo, Norway.
        {\tt\small tonnesfn@ifi.uio.no}}%
}

\begin{document}

\maketitle
\thispagestyle{empty}
\pagestyle{empty}


\begin{abstract}
  If robots are to become ubiquitous, they will need to be able to adapt to complex and dynamic environments.
  Robots that can adapt their bodies while deployed might be flexible and robust enough to meet this challenge.
  Previous work on dynamic robot morphology has focused on simulation, combining simple modules, or switching between locomotion modes.
  Here, we present an alternative approach: a self-reconfigurable morphology that allows a single four-legged robot to actively adapt the length of its legs to different environments.
  We report the design of our robot, as well as the results of a study that verifies the performance impact of self-reconfiguration.
  This study compares three different control and morphology pairs under different levels of servo supply voltage in the lab.
  We also performed preliminary tests in different uncontrolled outdoor environments to see if changes to the external environment supports our findings in the lab.
  Our results show better performance with an adaptable body, lending evidence to the value of self-reconfiguration for quadruped robots.
\end{abstract}


\section{INTRODUCTION}

Robots are increasingly asked to operate in more dynamic and unpredictable environments, alongside other robots, or humans.
The challenges of these environments can be handled through complex locomotion control or mechanical compliance~\cite{eckert_ICRA15}, or by giving a robot the ability to adapt and learn.
So far, robotic adaptation has focussed on a robot's control system, but adapting the body of a robot--its morphology--can provide a more fundamental flexibility~\cite{murata_RAM07}.
The concept of embodied cognition suggests that the interaction of the mind, body, and environment can all contribute to the task solving ability of the robotic system \cite{wilson_frontiers13_embodied}.
Earlier work in evolutionary robotics has also shown that different morphologies emerge for environments of varying complexity \cite{auerbach_plos14}.
Thus, a robot should adapt its body as well as its control system, to the environment and the task at hand.

\begin{figure}
  \centering
  \includegraphics[width=0.48\textwidth]{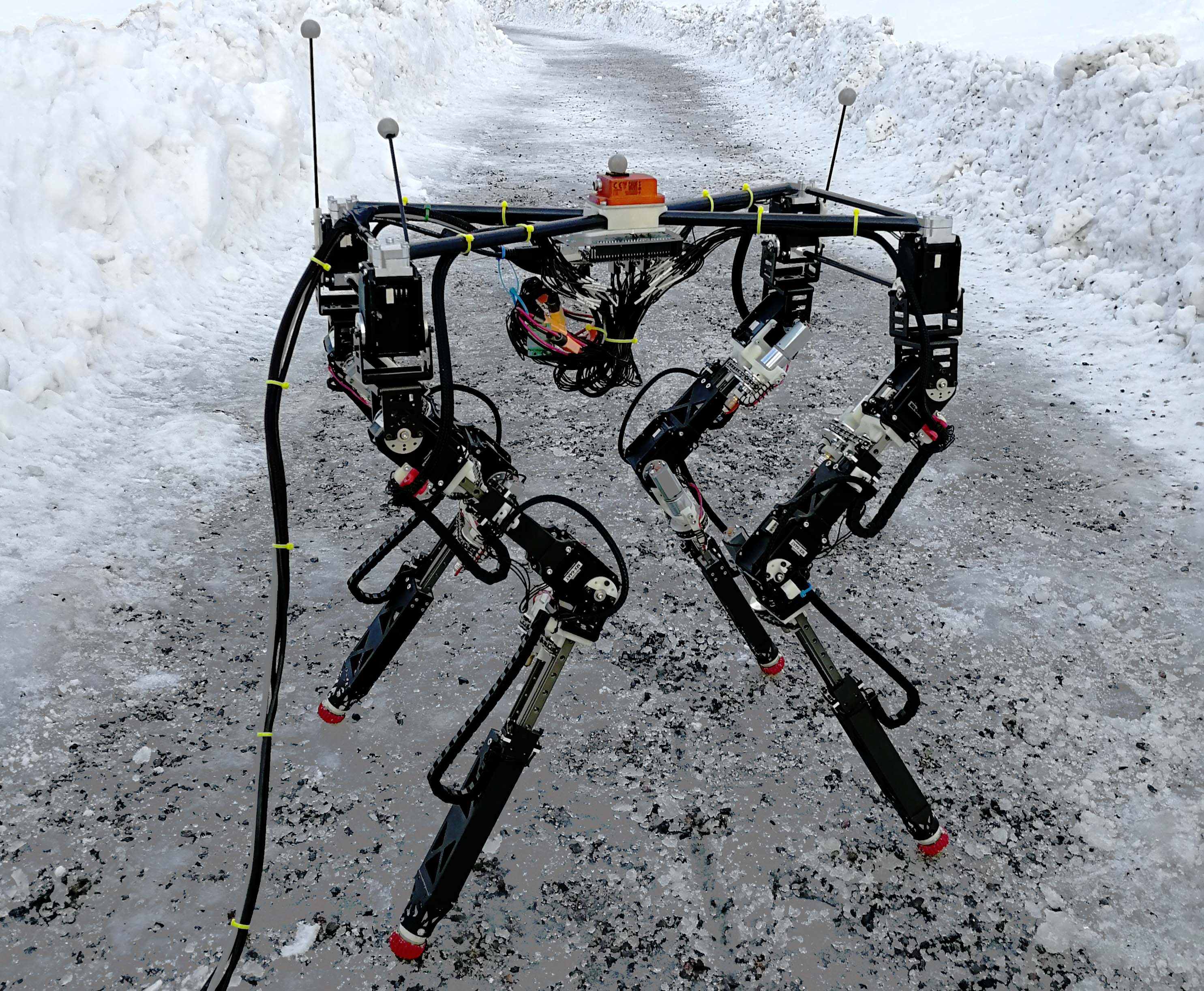}
  \caption{The robot during outdoor experiments. The legs can automatically change length, enabling experiments in self-reconfiguration.}
  \label{fig.robot}
  \vspace{-2.75mm}
\end{figure}

In this paper, we introduce a practical four-legged robot including self-reconfigurable legs (Fig.~\ref{fig.robot}).
Each leg features three rotational joints that are used for locomotion, and two prismatic joints.
While the prismatic joints are too slow to use in locomotion, they can actively alter the morphology of the robot during operation.
This ability might be applied to adapt to a dynamic environment; alternatively, the body can be changed during traditional single-morphology experiments to validate solutions on different robot bodies.

With this robot, we wish to investigate whether adapting morphology can lead to an advantage in tackling dynamic situations or environments.
This research goal can be summarised as the following hypothesis: \textit{No single robot morphology performs best for all situations, tasks or environments}.
That is, robots with dynamic morphology will be able to perform better than static morphologies, by modifying their own body in the face of changing situations.

In this paper, we present evidence supporting this hypothesis for our robot.
In the lab, we examined how different morphologies with hand-tuned gaits perform when the torque of the servos is changed.
This is done by constraining the robot's supply voltage, which emulates in-the-field depletion of a robot's battery.
We also describe preliminary tests of the robot under battery power in two field environments to explore our results in more difficult environmental conditions.
Our lab experiments support our hypothesis, and the field tests support our findings in the lab.

\paragraph*{Contribution}
There are two significant contributions in this paper.
First, we demonstrate and introduce a practical robot system for researching self-modifying morphology.
This system has been released as a fully certified open source hardware project, and can be used freely by other researchers.
Secondly, we show through experimental results--both from the lab, and from field scenarios--that having a self-reconfigurable morphology helps our robot to maintain optimal performance when adapting to changing supply voltages and external environments.
These experiments indicate that self-reconfigurable legs could improve the performance of robots doing complex tasks in dynamic environments.

\section{Background}

Being able to use different modes of locomotion will allow a robot to adapt to the most appropriate way of travel in dynamic and unknown environments.
Some robots are able to change their locomotion mode without morphological change \cite{Peterson_IROS11}, while others change it by switching between separate structures, such as wheels and flight rotors~\cite{kossett_ICRA10}.
Structures can also be shared by reusing parts of the body for different modes \cite{daler_IROS13}, which saves weight at the expense of mechanical and control complexity.
In general, these robots have discrete morphologies used for each mode of locomotion, precluding adaptation from a continuous range of different morphologies in response to internal or external factors.
Other robots can change parts of their bodies through adjustable compliance mechanisms~\cite{varimpact}, but these typically results in a much smaller impact on locomotion capability.

Modular self-reconfigurable robot systems, including complex simulations and physical implementations, can be divided into three architectures \cite{Yim_RAM07}.
The simplest architecture is the chain or tree architecture, with a serial connection between modules.
Zykov et al.~\cite{zykov_IROS08W} review many examples of systems that use this simple architecture, but still manage to show reasonably complex configurations.
The lattice architecture has modules connected in parallel along a two or three-dimensional grid.
This allows for more advanced base architectures, and connecting sub-parts of the system into meta-modules can yield interesting possibilities when changing the morphology of the system \cite{christensen_ICRA06}.
Other modular robots follow a mobile architecture, and can take on either of the previous architectures, or work as separate units.
This is closely related to the field of swarm robotics; physically connecting a swarm of robots to form new, cooperative morphologies can yield very flexible solutions \cite{mondada_IROS03}.
Despite these advances, modular robots still have a very coarse granularity when it comes to its morphology, when compared to other areas that change a robot's body.

The field of evolutionary robotics uses techniques from evolutionary computation to optimize control and--less often--morphology.
Evolution of both control and morphology together is usually performed in simulation and presents additional challenges due to the complex search-space~\cite{tonnesfn_evorobot17}.
The difference between performance in a simulator and a real-world counterpart is referred to as the \textit{reality gap}, and often makes it very challenging to transfer a result to the real world.
A lot of interesting research has been done in simulation alone, but there are many reasons to move more of the research into hardware, as described in one of three grand challenges to the ER field posed by Eiben \cite{eiben_frontiers14}.
There are some examples of evolution of morphology in hardware, but these require either excessive human intervention~\cite{milan17}, or use slow external reconfiguration of modular systems~\cite{vujovic2017evolutionary}.
There are also examples of self-reconfiguring morphology used exclusively to guide the search for a better controller of a single hand-designed optimal morphology ~\cite{Bongard11}.
In our previous work, we have demonstrated in that earlier versions of the DyRET platform, presented in this paper, can be used for evolutionary experiments to optimize morphology using mechanical self-reconfiguration~\cite{tonnesfn_GECCO18}.
Our work was the first example of such an approach as far as we are aware.

\section{System overview}

\begin{figure}
  \centering
  \begin{overpic}[width=0.495\textwidth]{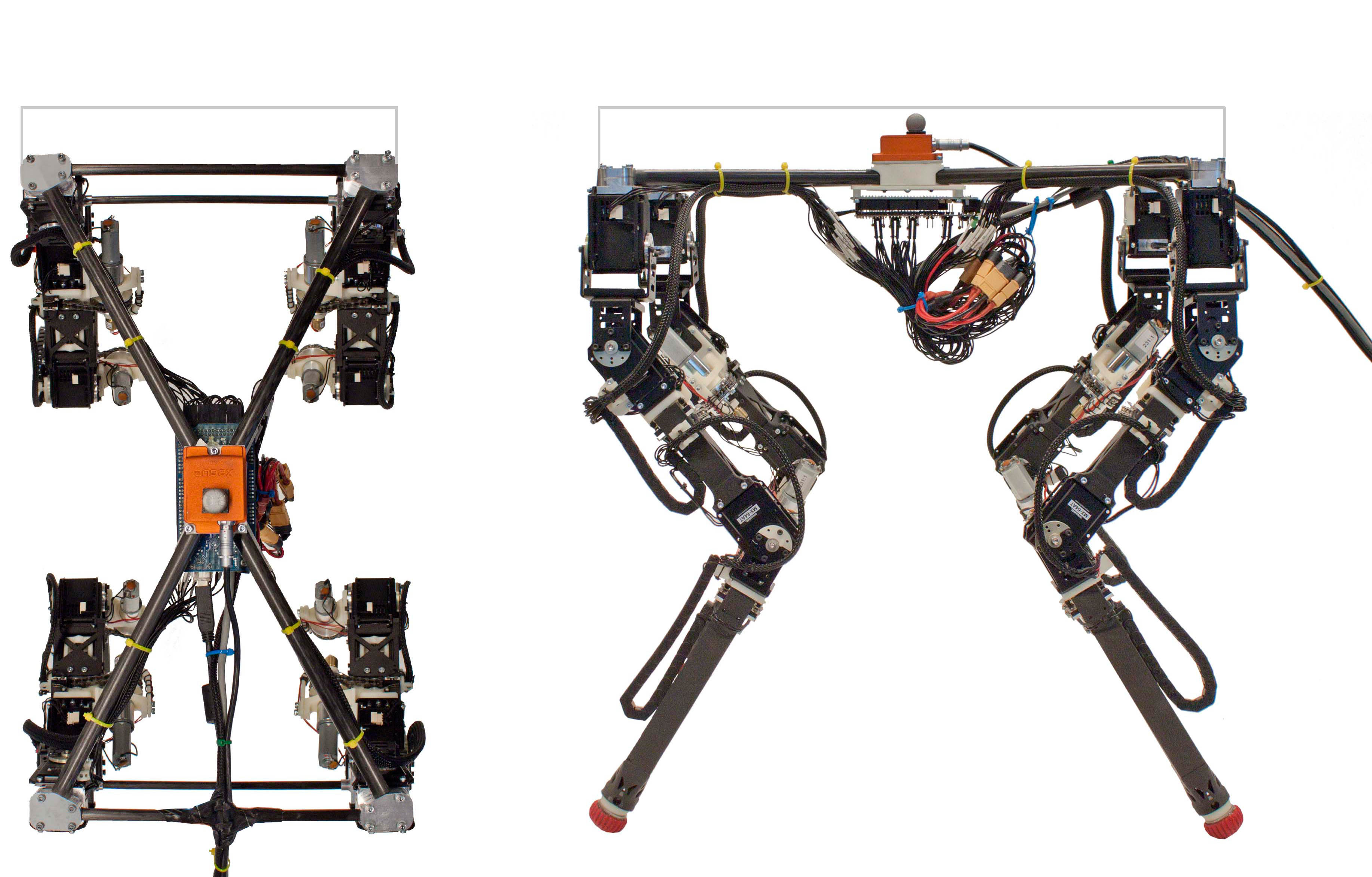}
  \put (62,58)  {470mm}
  \put (10.5,58)  {270mm}
  \end{overpic}
  \caption{Top and side view of the robot, with measurements. The height of the robot is dependent on pose and leg length, and is typically between 600mm and 700mm in normal operation.}
  \label{fig.dyret_diagram}
\end{figure}

\begin{figure}
  \centering
  \begin{overpic}[width=0.4\textwidth]{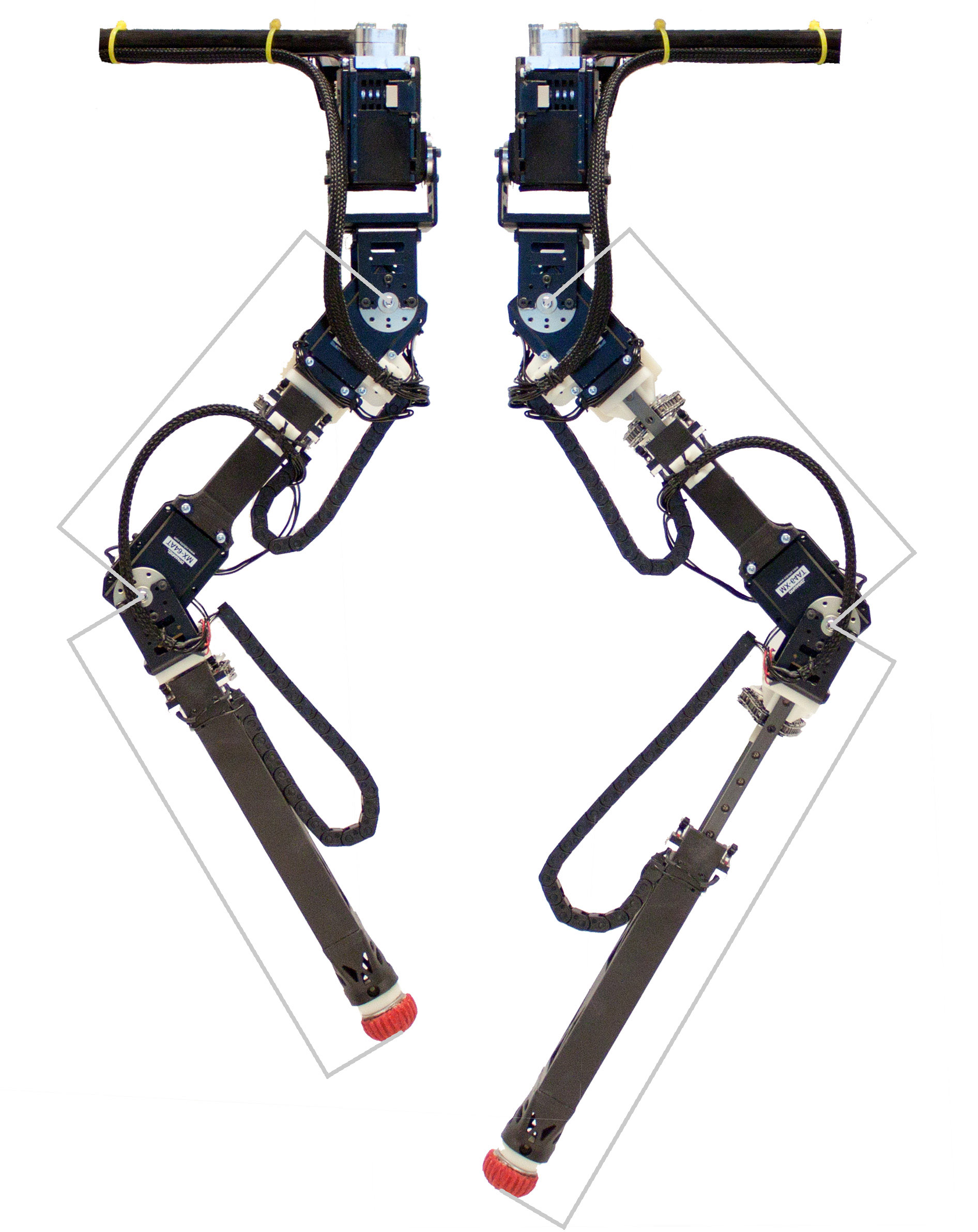}
  \put (33,60)  {Femur}
  \put (8,66)  {\rotatebox{50}{185mm}}
  \put (59.5,74)  {\rotatebox{-50}{205mm}}
  \put (8,35) {\rotatebox{-60}{255mm}}
  \put (59.5,20) {\rotatebox{60}{331mm}}
  \put (35,35) {Tibia}
  \put (44.5,67)   {\rotatebox{-52}{\begin{tikzpicture} \draw[red,ultra thick,dashed] (0,0) -- (1.2,0) -- (1.2,1) -- (0,1) -- (0,0); \end{tikzpicture}}}
  \end{overpic}
  \caption{Diagram of the legs in their rest pose, showing the two lengths used in the paper. The shortest available length is to the left, and 80\% of available length to the right, as used in our experiments. See Fig.~\ref{fig.reconfig_mechanism} for details on the reconfiguration mechanism in the red square.}
  \label{fig.leg_comparison}
\end{figure}
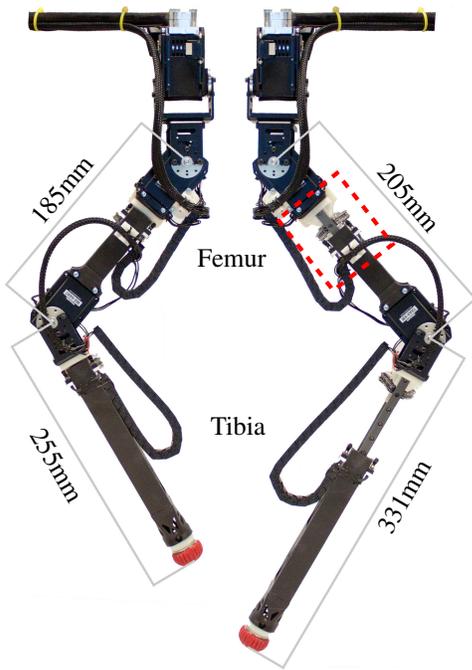

Our robot was developed to be a platform for experiments on self-adaptive morphologies and embodied cognition, and is shown in Fig.~\ref{fig.dyret_diagram}.
It is a certified open source hardware project, and documentation, code and design files are freely available online~\cite{github}.
It can actively reconfigure its morphology by changing the lengths of the two lower links of its legs, the femur and tibia.
The difference in height is illustrated in Fig.~\ref{fig.leg_comparison}.
Since it is used with machine learning techniques, the robot must withstand falls and unstable gaits, making maintainability and robustness important design factors.

Changing the length of the legs moves the center of gravity in the robot, affecting the balance.
Longer legs also mean lower servo rotational velocity for a given end-effector path, at the expense of higher torque requirements.
The length of the legs can therefore be used to mechanically gear the motors, and allow the robot to change where it sits in the trade-off between movement speed and force surplus continuously and autonomously.

\subsection{Mechanics}

The robot applies a mammal-inspired quadruped configuration.
All parts can either be bought as relatively inexpensive commercial off-the-shelf components, or be printed on consumer-grade 3D printers.
The parts for the robot without sensors are estimated to be about 6500USD in 2018, including 4300USD for the 12 servos alone.
Some parts can optionally be made in aluminium for improved robustness, which is relevant if the robot is used for gait learning experiments.
The main body of the robot is constructed with carbon fiber tubing of different diameters, which ensures a stable but low weight base for the four legs.
The complete robot weighs 5.5kg, and operates tethered during all experiments.

The robot has four legs with five degrees of freedom each.
The coxa (hip), femur (top leg), and tibia (lower leg) are all connected to revolute joints like traditional mammal robots, in addition to two prismatic joints to allow self-modification of the leg lengths.
Each leg includes three Dynamixel MX-64AT servos, with integrated PID controllers that receive angle commands over USB.
Off-the-shelf aluminium brackets are used to connect the servos to the rest of the robot where possible, with remaining connections using custom 3D printed and machined aluminium parts.

\begin{figure}
  \centering
  \includegraphics[width=0.26\textwidth]{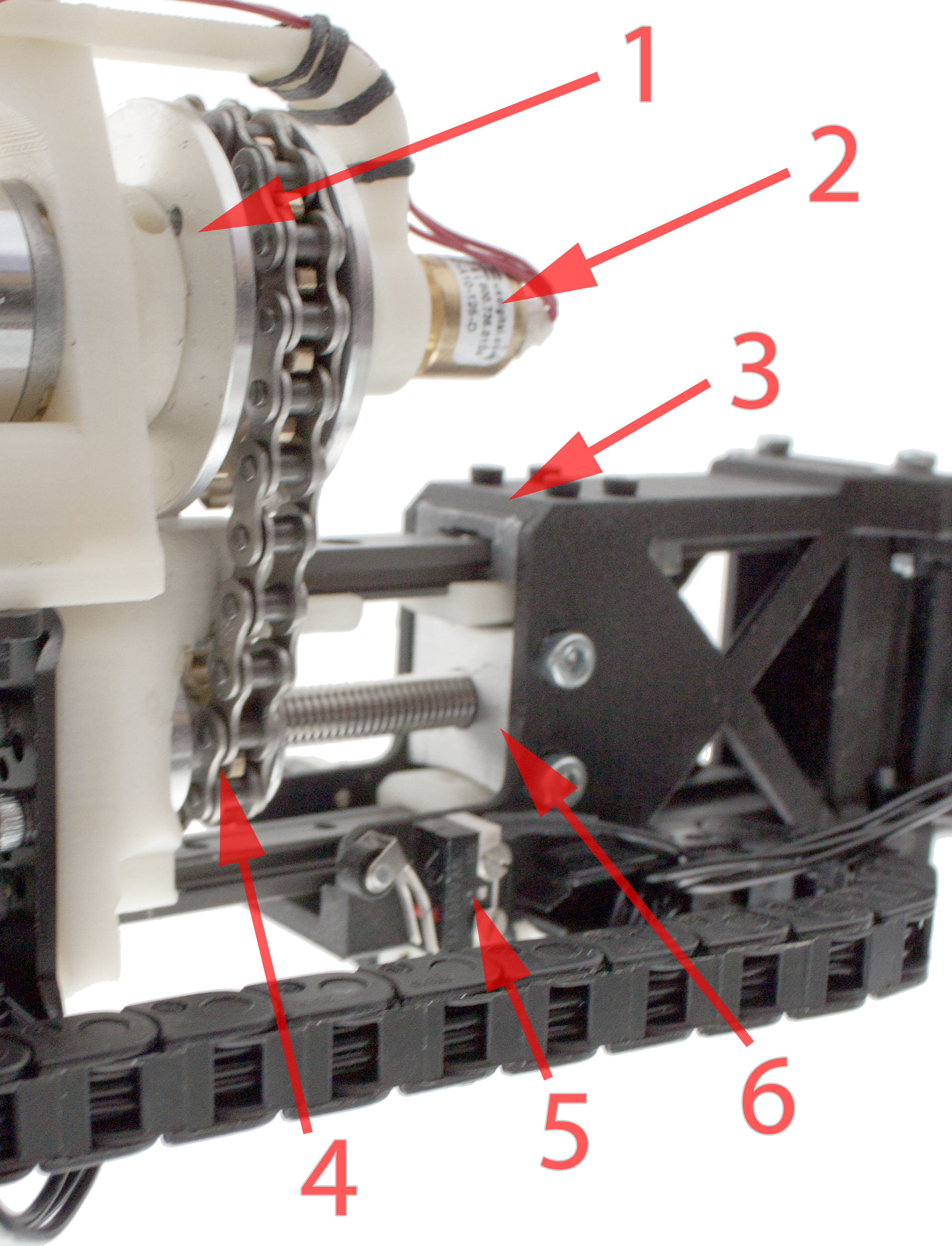}
  \caption{The reconfiguration mechanism, with the rest of the leg extending to the right of the image.
    (1) Brushed DC motor and sprocket,
    (2) Encoder for positioning,
    (3) Aluminium rail and carriages,
    (4) Threaded rod and sprocket,
    (5) Limit switch for zeroing,
    (6) Nut fixed to the movable part of the leg
  }
  \label{fig.reconfig_mechanism}
\end{figure}

The two lower links of each leg, femur and tibia, can be reconfigured to different lengths.
The reconfiguration mechanics is shown in Fig. \ref{fig.reconfig_mechanism}.
These linear actuators consist of small highly geared DC motors connected to lead screws through roller chain.
The leg is connected to the lead screw with a self-lubricating plastic nut, and rides on aluminium rails by two carriages.
The length of the leg is sensed by the encoder, which is calibrated on power-up using the mechanical end-stop.
Leg cabling has been run through cable carriers that keep the cable runs constant regardless of leg length, along with cable lacing techniques to secure the cables with minimal strain.
The low linear actuation speed ($\approx$1mm/s) makes it ineffective to use this mechanism actively during the gait, so it is exclusively used for changing the morphology configuration.

\subsection{Electronics}
All twelve servos are connected to an external computer running the software through a USB serial adapter.
Angular positions of all servos are reported to the system at approximately 60Hz, and new angle commands are received at the same rate.
Temperature, current and load are also read, to ensure the servos stay within operating specifications.
Servo position control is achieved using integrated PID-control in each servo.

The length of the reconfigurable legs are controlled using an Arduino Mega 2560 Rev 3 board with a custom PCB shield, which communicates with the software system through USB at 10hz.
Limit switches are routed directly to the digital inputs of the microcontroller with internal pull-ups, and all encoders are connected directly to the analog inputs.
The custom shield has twelve H-bridges to drive the DC motors in the linear actuators, which are controlled by PWM from the microcontroller.
Since we are using a screw mechanism for the linear actuators with inherently high holding load and friction, a proportional controller for each prismatic joint is sufficient to achieve stable actuation with 0.5mm accuracy.

An Xsens MTI-30 attitude and heading reference system (AHRS) is mounted close in the middle of the body to measure linear acceleration, rotational velocity and magnetic fields, giving data on absolute orientation at 100Hz.
Reflective markers are mounted on the main body of the robot to allow motion capture equipment to record the position and orientation of the robot at 100Hz.
The robot can carry enough weight to accommodate a full sensor package, such as a LIDAR and a depth camera, as well as an on-board Intel NUC computer.

\subsection{Software}

\begin{figure}
  \centering
  \includegraphics[width=0.45\textwidth]{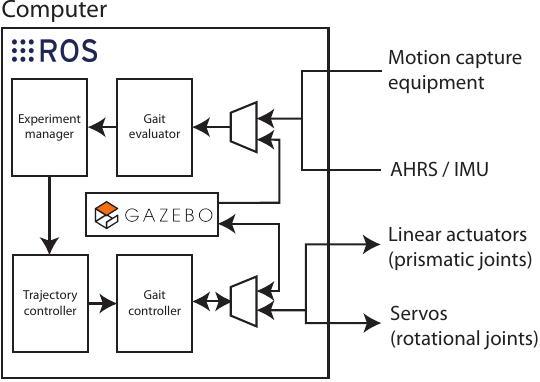}
  \caption{Overview of the software system. Each named black square is a node, and the system can either be connected to the Gazebo simulator, or to the physical robot and sensors in the real world.}
  \label{fig.software_diagram} 
\end{figure}

All software functions are implemented as separate Robot Operating System~\cite{ros} nodes in C++, and an overview of the system with its main nodes is shown in Fig~\ref{fig.software_diagram}.
An experiment manager node takes input from the user, and runs the different experiments.
Trajectories with distance to move, direction, and configuration are sent to a trajectory controller that interfaces to the gait controller.
Several different gait controllers can be used, as switching out nodes are simple plug-and-play procedures that can be done during system operation.
The gait controller either sends commands to the hardware in the real world, or to the Gazebo simulator \cite{koenig_IROS04_gazebo}.
Feedback on performance is received by the gait evaluator from either simulations or the real world, and is analyzed, logged, and reported back to the experiment manager node.

\subsection{Control}

We have successfully implemented and used both high-level and low-level gait control.
Only the high-level control is used for experiments with self-reconfiguration, as the gaits produced are more robust and easier to change for an engineer than low-level gaits.
Low-level control is detailed in~\cite{Nordmoen18}, and not used for the experiments in this paper.

The high-level control is an inverse-kinematics based position controller for the legs of the robot, making it easy for an engineer to hand design a gait, as well as to intuitively understand gaits that have been optimized by machine learning algorithms.
It generates a continuous, regular crawl gait, and the body moves at a constant forward speed during the gait sequence, lifting each leg separately to maximize stability.
The gait controller uses 8 parameters to generate points along an interpolating looping cubic hermite spline, seen in table \ref{table.handParams}.
Ground height is given in millimetres ($H_{\text{ground}}$), and is dependent on the femur length ($L_{\text{femur}}$) and tibia length ($L_{\text{tibia}}$), following this equation:

\begin{equation} \label{eq.groundheight}
\begin{split}
H_{\text{ground}} = -430 - ((L_{\text{femur}} + L_{\text{tibia}}) * 0.8);
\end{split}
\end{equation}

The 5 control points (x,y,z) for the spline are derived from step length ($L_{\text{step}}$), step height ($H_{\text{step}}$), step smoothing ($S$), and ground height ($H_{\text{ground}}$):
\begingroup\makeatletter\def\f@size{6}\check@mathfonts
\begin{equation}
\begin{split}
  &(0,     \frac{L_{\text{step}}}{2},                               H_{\text{ground}})\\  
  &(0,    \frac{-L_{\text{step}}}{2},                               H_{\text{ground}})\\  
  &(0,    \frac{-L_{\text{step}}}{2}, H_{\text{ground}} + \frac{H_{\text{step}}}{1.5})\\  
  &(0,                             0,             H_{\text{ground}} + H_{\text{step}})\\  
  &(0, \frac{L_{\text{step}}}{2} + S,   H_{\text{ground}} + \frac{H_{\text{step}}}{4})    
\end{split}
\label{eq.points}
\end{equation}
\endgroup

A balancing wag movement ($W$) is added to allow the robot to lean to the opposite side of the leg it is currently lifting to allow for statically stable gaits.
This is added to the position from the spline at each time step ($t$).
Period ($T$) is calculated from the gait frequency parameter ($f$), while phase offset ($W_\phi$) and amplitudes ($W_\text{Ax}$ and $W_\text{Ay}$) comes directly from the gait parameters.
An offset of 0.43 is added to offset forward and sideways movement.
\begingroup\makeatletter\def\f@size{6}\check@mathfonts
\begin{equation}
\begin{split}
&W_x = \frac{A_\text{x}}{2} * tanh(3*sin( \frac{2\pi*(t+(W_\phi*T))}{T}))\\
&W_y = \frac{A_\text{y}}{2} * tanh(3*sin( \frac{2\pi*(t+(W_\phi+0.43)*\frac{T}{2})}{\frac{T}{2}}))
\end{split}
\end{equation}
\endgroup

We have also added a parameter for a lift duration ($D_\text{lift}$) to control what percentage of the gait period is used to lift the leg back to the front.

\section{Experiments and results}

We tested DyRET to find whether changing morphology really can deliver an advantage in different environments.
We designed experiments in different environments that we speculated would require different leg lengths.
In these limited number of environments, our hypothesis can be falsified if one morphology is the best performer in all situations.
There are many measures of performance that could be used, but we chose to only look at forward speed, as it is both simple to measure and understand.
By studying how different morphologies impact performance, we can gain an indication of whether mechanical self-reconfiguration could be useful in real-world dynamic environments.

Our experimental procedure was as follows.
First, we hand designed the gaits to be used in the experiments by choosing conservative gait parameters.
The main experiment is done in the lab to investigate how reducing the supply voltage, and thereby the torque of the motors, affects the performance of two different morphologies.
The lab environment gives us stable evaluations, and the results should be directly transferable to real-world applications where the servos run on unregulated battery power.
Then, to evaluate the plausibility of the lab experiments, we perform preliminary evaluation of the gaits in two different field environments: an indoor garage facility, and an outdoor footpath in winter conditions.

We used two morphology configurations for the experiments: one that uses the shortest available leg length (referred to as "short robot"), and one that uses 80\% of the available length (referred to as "tall robot").
The maximum length of the legs was designed with optimal lab conditions in mind, so we use only 80\% of the available length of both links as to not strain the robot in the more demanding environments.

\begin{figure}
  \centering
  \includegraphics[width=0.40\textwidth]{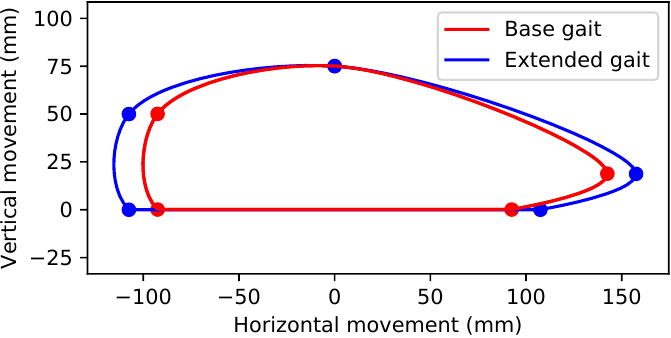}
  \caption{The leg path and control points for the gaits used in our experiments.}
  \label{fig.splines}
\end{figure}

We hand designed parameters for a conservative gait (referred to as "base gait") based on experience from previous experiments \cite{tonnesfn_GECCO18,tonnesfn_ICES16}.
The walking speed of the robot is limited by the maximum rotational speed of the servos.
This speed is a function of torque, but we have selected a maximum allowable rotational speed of 25RPM in our current setup, based on specifications and experience.
Since the legs of the tall robot are longer, the same leg endpoint movement requires a smaller rotational change.
This means that the taller robot can walk faster than the shorter robot, given the same rotational speed limit.
We therefore included a faster gait (referred to as "the extended gait") that could only be used on the taller robot without exceeding servo specifications.
This gait has both increased frequency and step length, while all other gait parameters are kept the same, as seen in Table~\ref{table.handParams}.
The spline path for each gait can be seen in Fig.~\ref{fig.splines}.
The base gait is evaluated on both morphologies, while the extended gait is only valid for the tall robot.

\begin{table}
  \centering
  \caption{Hand designed gait parameters}
  \label{table.handParams}
  \begin{tabular}{  l  c  r  r }
    \hline
    \bfseries Parameter & \bfseries Symbol  & \bfseries Base gait & \bfseries Extended gait \\
    \hline
    Step length         & $L_{\text{step}}$ & 185mm               & 215mm                   \\
    Step height         & $H_{\text{step}}$ & 75mm                & 75mm                    \\
    Smoothing           & $S$               & 50mm                & 50mm                    \\
    Frequency           & $f$               & 0.275hz             & 0.35hz                  \\
    Lift duration       & $D_\text{lift}$   & 20\%                & 20\%                    \\
    Wag phase           & $W_\phi$          & 0.0                 & 0.0                     \\
    Wag amplitude x     & $W_\text{Ax}$     & 15mm                & 15mm                    \\
    Wag amplitude y     & $W_\text{Ay}$     & 10mm                & 10mm                    \\
    \hline
  \end{tabular}
  \vspace{-3mm}
\end{table}

\subsection{Lab experiments}
In the lab experiments, we change the supply voltage of the servos to investigate whether different leg lengths are needed when the torque available to the robot changes.
Each gait and morphology pair was evaluated ten times with two different supply voltages by the robot walking 1.5m forwards and then 1.5m in reverse.
This takes up to about 60s, depending on the gait speed.
Pair-wise Mann-Whitney U tests with Holm correction were performed to assess statistical significance of differences.

\begin{figure}
  \centering
  \includegraphics[width=0.46\textwidth]{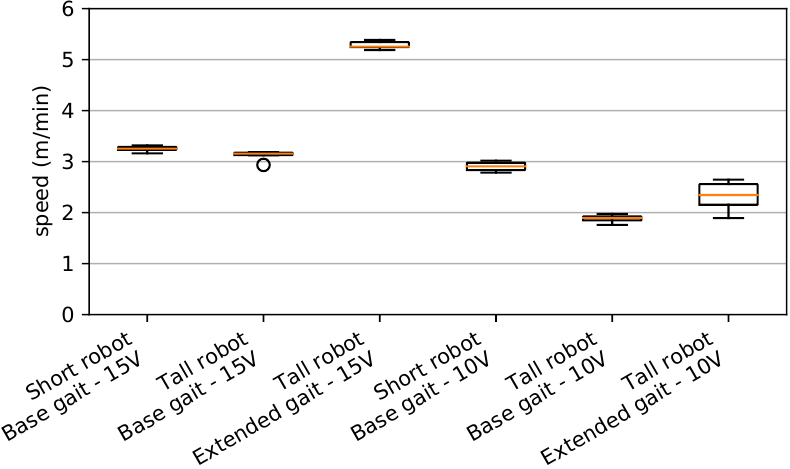}
  \caption{Results of the lab evaluations of the three different gait and morphology pairs at the different supply voltages.}
  \label{fig.boxplot_lab}
  \vspace{-4mm}
\end{figure}

The results are shown in Fig.~\ref{fig.boxplot_lab}, and with more details in table \ref{table.expLab}.
Using the base gait at the higher voltage, we see that the short and tall morphology perform similarly at just over 3m/min.
The tall robot with extended gait (Mdn $=5.26$) perform significantly better than both the short robot (Mdn $=3.26$, U $=0$, p $<0.001$), and the tall robot with the base gait (Mdn $=3.16$, U $=0$, p $<0.001$).

At the lower voltage, we see that the short robot has a slight decrease in performance to just under 3m/min, while the tall robot with the base gait is now unable to match that speed, with a reduction to below 2m/min.
The short robot (Mdn $=2.90$) now outperforms the tall robot with both base gait (Mdn $=1.90$, U $=0$, p $<0.001$) and extended gait (Mdn $=2.35$, U $=0$, p $<0.001$).

For the lab experiments, we see that at the high voltage, the tall morphology performs best, with its extended gait.
At the lower voltage, the short morphology performs best.

\subsection{Field experiments}

The field experiments in the garage and on the footpath, involved changes in the external environment, including surface friction, texture, temperature and humidity, to see if this affects the performance of different morphologies.
In both field environments, the robot was powered by an external three-cell LiPo battery pack (11.1V) and controlled from a tethered laptop.
The garage environment had a smooth concrete floor, with much lower friction than the lab's carpet, and ambient temperature of around $+4^{\circ}$C.
The outdoor evaluations were held on a footpath in Norwegian winter conditions (around  $-5^{\circ}$C) where the surface was a mix of compacted snow, ice, and gravel -- a very challenging environment to retain traction, shown in Fig.~\ref{fig.robot}.
Three combinations of morphology and gait were evaluated by 120 seconds of forward walking in each environment.
Each evaluation was replicated twice in the garage and three times on the outdoor footpath.
Speed was evaluated by using a hand-held laser distance measurer along with a time measurement from the test program.
The field experiments are only done as a preliminary investigation to see if our lab experiments are feasible also for external environment changes.
We therefore have a limited number of evaluations, and are not able to do Mann-Whitney U tests to analyze statistical significance.

\begin{figure}
  \centering
  \includegraphics[width=0.48\textwidth]{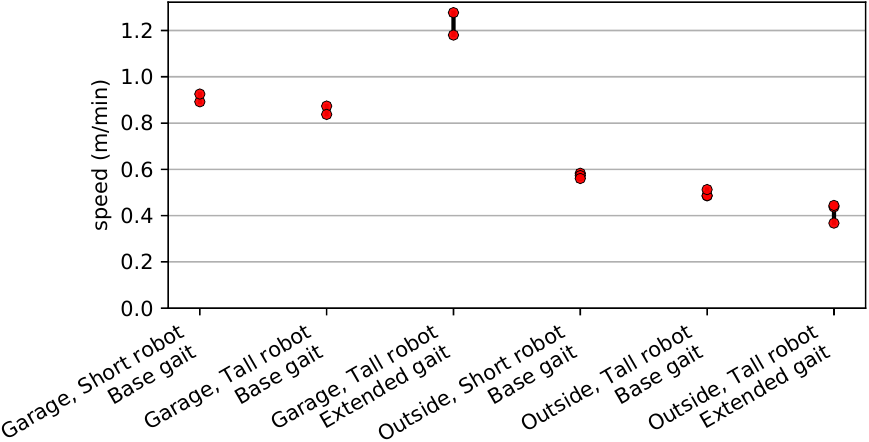}
  \caption{Results from the field experiments. Note that each garage evaluation has two data points, while the outdoor evaluations have three.}
  \label{fig.boxplot_rw}
\end{figure}

The results from the field experiments are shown in Fig.~\ref{fig.boxplot_rw}, with details in table \ref{table.expLab}.
Results from the garage show a big reduction in performance from the lab; the base gaits achieve a speed of about 0.9m/min, while the tall robot with the extended gait achieves a speed of about 1.2m/min.
Although there is a reduction in speed when compared to lab conditions, we see the same trend as with the high voltage experiment in the lab: both morphologies perform similarly using the base gait, while the tall morphology with the extended gait walks faster.

On the footpath, all speeds are further reduced; and we see the same trend as observed with the lower voltage experiment in the lab.
The shorter robot now outperforms the tall robot with both the extended gait, as well as base gait.

Our field experiments indicate that for the less demanding garage environment, the tall morphology performs best with its extended gait.
In the more demanding outdoor environment, the short morphology performs best.

\begin{table}
  \centering
  \caption{Experiment results details}
  \label{table.expLab}
  \scalebox{0.9}{
  \begin{tabular}{  c | c  c  c  c  c  }
    \hline
    \bfseries Experiment & \bfseries Evals & \bfseries Morphology & \bfseries Gait & \bfseries Range & \bfseries Mean \\
    \hline
    \multirow{3}{*}{Lab, 15V} & 10 & Short & Base     & [3.161, 3.319] & 3.252 \\
                              & 10 &  Tall & Base     & [2.936, 3.186] & 3.135 \\
                              & 10 &  Tall & Extended & [5.188, 5.388] & 5.283 \\
    \hline
    \multirow{3}{*}{Lab, 10V} & 10 & Short & Base     & [2.784, 3.019] & 2.902 \\
                              & 10 &  Tall & Base     & [1.759, 1.977] & 1.884 \\
                              & 10 &  Tall & Extended & [1.893, 2.647] & 2.326 \\
    \hline
    \multirow{3}{*}{Garage}   & 2  & Short & Base     & [0.892, 0.926] & 0.909 \\
                              & 2  &  Tall & Base     & [0.838, 0.874] & 0.856 \\
                              & 2  &  Tall & Extended & [1.181, 1.278] & 1.229 \\
    \hline
    \multirow{3}{*}{Outside}  & 3  & Short & Base     & [0.561, 0.583] & 0.572 \\
                              & 3  &  Tall & Base     & [0.486, 0,513] & 0.495 \\
                              & 3  &  Tall & Extended & [0.368, 0.444] & 0.416 \\
    \hline
  \end{tabular}
  }
\end{table}

\section{Discussion}

The fact that the tall robot performed best at high supply voltage, and the short robot performed best at lower supply voltage, supports our hypothesis in lab conditions.
This also strengthens our assumption that the self-adaptive legs can be used to adapt the robot to the new supply voltage by selecting different trade-offs between speed and torque.

In our preliminary indoor field experiments, we observed that the tall robot with the extended gait outperformed the short robot.
In the more demanding outdoor environment, we saw that the short robot now outperformed the tall robot with either gait.
This suggests that the trade-off between speed and torque in the shorter robot suits this new and demanding environment better than the taller morphology.
The field experiments supports our findings in the lab, and suggests further exploration could be beneficial.

Our field experiments in Fig.~\ref{fig.boxplot_rw} showed a reduced performance for all individuals when compared to the same gaits in lab conditions in Fig.~\ref{fig.boxplot_lab}.
A different surface, temperature, control computer, and running on battery are all factors that could have contributed to this, and we observed slipping and stumbling of the robot that we had not previously seen in the lab.

The fact that we are working on a physical robot system has severely limited the number of samples we have used in our experiments.
It is challenging to do longer experiments in outdoor environments, where there is a large number of variables that can not be controlled compared with lab experiments.
The observed variance, however, was quite low, and statistical significance has been assessed where possible to help support our conclusions.

We are only able to address the hypothesis for our own robot, and the specific morphologies and environments in our experiments.
We believe the main reason for the observed difference in performance between voltages or environments comes from the leg length gearing the motors and allowing different speed/torque tradeoffs.
This effect would act linearly, and we do not expect to find new morphologies that outperform those we used in all our selected environments.
Our results are encouraging, and suggest that other robots with self-reconfigurable hardware might derive similar advantages from adapting their bodies as we have.

\section{Conclusion and future work}

\addtolength{\textheight}{-110mm} 

In this paper we introduced a novel four-legged mammal-inspired robot with mechanical self-modifying morphology.
We hypothesise that no single robot morphology performs best for all situations, tasks or environments.
To address this for our robot we ran lab experiments showing that different servo torques require different morphologies to perform well.
We also performed preliminary field testing of the robot in two outdoor environments, which supported the results of our lab experiments.
These results indicate that mechanically self-modifying robots may perform better in dynamic environments by adapting morphology as well as control to new conditions.

Even though we have shown clear indications that different morphologies are optimal in different situations or environments, we have yet to investigate how to switch between these or utilize a library of gaits and morphologies efficiently and autonomously.
Doing more extensive field experiments with more evaluations and better tailored test setups would allow investigating how to automatically and continuously adapt control and morphology.
We would also be able to investigate optimal morphologies for different environments, and the close ties and interactions between the environment and a robot's control and morphology.
We evaluated the robot in two different outdoor environments, one of which was very challenging for the robot.
It would be interesting to do more realistic experiments in a more extensive collection of environments, and introduce dynamic elements such as other robots or humans that might also affect the efficiency of different morphology-controller pairs.

We also hope that these experiments inspire more research on real world mechanical reconfiguration, and that our newly developed and open sourced platform might help lower the initial investment needed to begin such research by allowing others to use or extend our robot design~\cite{github}, either in simulation or the real world.

\bibliographystyle{IEEEtran}
\bibliography{bibliography}

\begin{thebibliography}{10}
\providecommand{\url}[1]{#1}
\csname url@samestyle\endcsname
\providecommand{\newblock}{\relax}
\providecommand{\bibinfo}[2]{#2}
\providecommand{\BIBentrySTDinterwordspacing}{\spaceskip=0pt\relax}
\providecommand{\BIBentryALTinterwordstretchfactor}{4}
\providecommand{\BIBentryALTinterwordspacing}{\spaceskip=\fontdimen2\font plus
\BIBentryALTinterwordstretchfactor\fontdimen3\font minus
  \fontdimen4\font\relax}
\providecommand{\BIBforeignlanguage}[2]{{%
\expandafter\ifx\csname l@#1\endcsname\relax
\typeout{** WARNING: IEEEtran.bst: No hyphenation pattern has been}%
\typeout{** loaded for the language `#1'. Using the pattern for}%
\typeout{** the default language instead.}%
\else
\language=\csname l@#1\endcsname
\fi
#2}}
\providecommand{\BIBdecl}{\relax}
\BIBdecl

\bibitem{eckert_ICRA15}
P.~Eckert, A.~Spröwitz, H.~Witte, and A.~J. Ijspeert, ``Comparing the effect
  of different spine and leg designs for a small bounding quadruped robot,'' in
  \emph{2015 IEEE International Conference on Robotics and Automation
  (ICRA15)}, May 2015, pp. 3128--3133.

\bibitem{murata_RAM07}
S.~Murata and H.~Kurokawa, ``Self-reconfigurable robots,'' \emph{IEEE Robotics
  Automation Magazine}, vol.~14, no.~1, pp. 71--78, March 2007.

\bibitem{wilson_frontiers13_embodied}
A.~Wilson and S.~Golonka, ``Embodied cognition is not what you think it is,''
  \emph{Frontiers in Psychology}, vol.~4, p.~58, 2013.

\bibitem{auerbach_plos14}
J.~E. Auerbach and J.~C. Bongard, ``Environmental influence on the evolution of
  morphological complexity in machines,'' \emph{PLoS computational biology},
  vol.~10, no.~1, 2014.

\bibitem{Peterson_IROS11}
K.~Peterson and R.~S. Fearing, ``Experimental dynamics of wing assisted running
  for a bipedal ornithopter,'' in \emph{2011 IEEE/RSJ International Conference
  on Intelligent Robots and Systems (IROS11)}, 2011.

\bibitem{kossett_ICRA10}
A.~Kossett, R.~D'Sa, J.~Purvey, and N.~Papanikolopoulos, ``Design of an
  improved land/air miniature robot,'' in \emph{2010 IEEE International
  Conference on Robotics and Automation (ICRA10)}, May 2010.

\bibitem{daler_IROS13}
L.~Daler, J.~Lecoeur, P.~B. Hahlen, and D.~Floreano, ``A flying robot with
  adaptive morphology for multi-modal locomotion,'' in \emph{2013 IEEE/RSJ
  International Conference on Intelligent Robots and Systems (IROS13)}, Nov
  2013, pp. 1361--1366.

\bibitem{varimpact}
B.~Vanderborght, A.~Albu-Schaeffer, A.~Bicchi, E.~Burdet, D.~Caldwell,
  R.~Carloni, M.~Catalano, O.~Eiberger, W.~Friedl, G.~Ganesh, M.~Garabini,
  M.~Grebenstein, G.~Grioli, S.~Haddadin, H.~Hoppner, A.~Jafari, M.~Laffranchi,
  D.~Lefeber, F.~Petit, S.~Stramigioli, N.~Tsagarakis, M.~V. Damme, R.~V. Ham,
  L.~Visser, and S.~Wolf, ``Variable impedance actuators: A review,''
  \emph{Robotics and Autonomous Systems}, vol.~61, no.~12, pp. 1601 -- 1614,
  2013.

\bibitem{Yim_RAM07}
M.~Yim, W.~min Shen, B.~Salemi, D.~Rus, M.~Moll, H.~Lipson, E.~Klavins, and
  G.~S. Chirikjian, ``Modular self-reconfigurable robot systems [grand
  challenges of robotics],'' \emph{IEEE Robotics Automation Magazine}, vol.~14,
  no.~1, pp. 43--52, March 2007.

\bibitem{zykov_IROS08W}
V.~Zykov, W.~Phelps, N.~Lassabe, and H.~Lipson, ``Molecubes extended:
  Diversifying capabilities of open-source modular robotics,'' in
  \emph{IROS-2008 Self-Reconfigurable Robotics Workshop}, 2008, pp. 22--26.

\bibitem{christensen_ICRA06}
D.~J. Christensen and K.~Stoy, ``Selecting a meta-module to shape-change the
  atron self-reconfigurable robot,'' in \emph{2006 IEEE International
  Conference on Robotics and Automation (ICRA06)}, May 2006.

\bibitem{mondada_IROS03}
F.~Mondada, A.~Guignard, M.~Bonani, D.~Bar, M.~Lauria, and D.~Floreano,
  ``Swarm-bot: from concept to implementation,'' in \emph{2003 IEEE/RSJ
  International Conference on Intelligent Robots and Systems (IROS03)}, Oct
  2003.

\bibitem{tonnesfn_evorobot17}
T.~F. Nygaard, E.~Samuelsen, and K.~Glette, ``Overcoming initial convergence in
  multi-objective evolution of robot control and morphology using a two-phase
  approach,'' in \emph{2017 Applications of Evolutionary Computation
  (EvoStar17)}, G.~Squillero and K.~Sim, Eds.\hskip 1em plus 0.5em minus
  0.4em\relax Springer International Publishing, 2017, pp. 825--836.

\bibitem{eiben_frontiers14}
A.~E. Eiben, ``Grand challenges for evolutionary robotics,'' \emph{Frontiers in
  Robotics and AI}, vol.~1, p.~4, 2014.

\bibitem{milan17}
M.~Jelisavcic, M.~de~Carlo, E.~Hupkes, P.~Eustratiadis, J.~Orlowski,
  E.~Haasdijk, J.~E. Auerbach, and A.~E. Eiben, ``Real-world evolution of robot
  morphologies: A proof of concept,'' \emph{Artificial Life}, vol.~23, no.~2,
  pp. 206--235, 2017.

\bibitem{vujovic2017evolutionary}
V.~Vujovic, A.~Rosendo, L.~Brodbeck, and F.~Iida, ``Evolutionary developmental
  robotics: Improving morphology and control of physical robots,''
  \emph{Artificial Life}, 2017.

\bibitem{Bongard11}
J.~Bongard, ``Morphological change in machines accelerates the evolution of
  robust behavior,'' \emph{Proceedings of the National Academy of Sciences},
  vol. 108, no.~4, pp. 1234--1239, 2011.

\bibitem{tonnesfn_GECCO18}
T.~F. Nygaard, C.~P. Martin, E.~Samuelsen, J.~Torresen, and K.~Glette,
  ``Real-world evolution adapts robot morphology and control to hardware
  limitations,'' in \emph{2018 Genetic and Evolutionary Computation Conference
  (GECCO18)}.\hskip 1em plus 0.5em minus 0.4em\relax ACM, 2018, pp. 125--132.

\bibitem{github}
\BIBentryALTinterwordspacing
T.~F. Nygaard, ``{DyRET GitHub repository}.'' [Online]. Available:
  \url{https://github.com/dyret-robot/dyret\_documentation}
\BIBentrySTDinterwordspacing

\bibitem{ros}
M.~Quigley, K.~Conley, B.~Gerkey, J.~Faust, T.~Foote, J.~Leibs, R.~Wheeler, and
  A.~Y. Ng, ``Ros: an open-source robot operating system,'' in \emph{ICRA
  workshop on open source software}, vol.~3, no. 3.2.\hskip 1em plus 0.5em
  minus 0.4em\relax Kobe, Japan, 2009, p.~5.

\bibitem{koenig_IROS04_gazebo}
N.~Koenig and A.~Howard, ``Design and use paradigms for gazebo, an open-source
  multi-robot simulator,'' in \emph{2004 IEEE/RSJ International Conference on
  Intelligent Robots and Systems (IROS04)}, vol.~3, Sept 2004, pp. 2149--2154
  vol.3.

\bibitem{Nordmoen18}
J.~Nordmoen, K.~O. Ellefsen, and K.~Glette, ``Combining map-elites and
  incremental evolution to generate gaits for a mammalian quadruped robot,'' in
  \emph{Applications of Evolutionary Computation}, K.~Sim and P.~Kaufmann,
  Eds.\hskip 1em plus 0.5em minus 0.4em\relax Cham: Springer International
  Publishing, 2018, pp. 719--733.

\bibitem{tonnesfn_ICES16}
T.~F. Nygaard, J.~Torresen, and K.~Glette, ``Multi-objective evolution of fast
  and stable gaits on a physical quadruped robotic platform,'' in \emph{2016
  IEEE Symposium Series on Computational Intelligence (SSCI)}, 2016, pp. 1--8.

\end{thebibliography}

\end{document}